\documentclass[letterpaper, 10 pt, conference]{ieeeconf}  

\IEEEoverridecommandlockouts                              

\usepackage{geometry}
\usepackage[utf8]{inputenc}

\usepackage{algorithm} 
\usepackage{program} 
\usepackage{algorithmic}
\usepackage{listings}
\usepackage{graphicx}
\usepackage{wrapfig}
\usepackage{comment}
\usepackage{hyperref}
\usepackage{listings}
\usepackage{caption}
\usepackage{subcaption}
\usepackage{xcolor}
\usepackage{lipsum}
\usepackage{placeins} 
\usepackage{color, colortbl}
\usepackage{pdfpages}

\newcommand\todo[1]{\textcolor{red}{#1}}

\renewcommand{\footnotesize}{\scriptsize}

\usepackage{caption,setspace}
\DeclareCaptionFormat{listing}{\rule{\dimexpr0.9\columnwidth+17pt\relax}{0.4pt}\par\vskip1pt#1#2#3}
\newcounter{code}



\definecolor{redalias}{HTML}{CA1236}

\title{
\emph{alurity}, a toolbox for robot cybersecurity
}

\author{
    Víctor Mayoral-Vilches$^{1,}$$^{2}$, Irati Abad-Fernández$^{2}$, Martin Pinzger$^{1}$, Stefan Rass$^{1}$, \\
    Bernhard Dieber$^{3}$, Alcino Cunha$^{4}$, Francisco J. Rodríguez-Lera$^{5}$, Giovanni Lacava$^{6}$, \\
     Angelica Marotta$^{6}$, Fabio Martinelli$^{6}$ and Endika Gil-Uriarte$^{2}$ 
\thanks{$^{1}$Víctor Mayoral-Vilches and Stefan Rass are with the System Security (SYSSEC) group. Martin Pinzger is with the Software Engineering group. All from Universität Klagenfurt, Austria.
        {\tt\footnotesize v1mayoralv@edu.aau.at}}%
\thanks{$^{2}$Víctor Mayoral-Vilches, Irati Abad-Fernández and Endika Gil-Uriarte are with Alias Robotics,  Spain.}%
\thanks{$^{3}$Bernhard Dieber is with the Institute for Robotics and Mechatronics, Joanneum Research, Austria.}%
\thanks{$^{4}$Alcino Cunha is with the Department of Informatics of University of Minho, Portugal. He's also with the High-Assurance Software Laboratory of University of Minho and with the INESC TEC associate laboratory.}%
\thanks{$^{5}$Francisco J. Rodríguez-Lera is with the Robotics Group at University of León, Spain.}%
\thanks{$^{6}$Giovanni Lacava, Angelica Marotta and Fabio Martinelli are with Istituto di Informatica e Telematica, Consiglio Nazionale delle Ricerche (CNR), Italy.}%
}

\begin{document}



\maketitle
\pagestyle{plain}

\begin{abstract}

The reuse of technologies and inherent complexity of most robotic systems is increasingly leading to robots with wide attack surfaces and a variety of potential vulnerabilities. Given their growing presence in public environments, security research is increasingly becoming more important than in any other area, specially due to the safety implications that robot vulnerabilities could cause on humans. We argue that security triage in robotics is still immature and that new tools must be developed to accelerate the \emph{testing-triage-exploitation} cycle, necessary for prioritizing and accelerating the mitigation of flaws.

The present work tackles the current lack of offensive cybersecurity research in robotics by presenting a toolbox and the results obtained with it through several use cases conducted over a year period. We propose a modular and composable toolbox for robot cybersecurity: \emph{alurity}. By ensuring that both roboticists and security researchers working on a project have a common, consistent and easily reproducible development environment, \emph{alurity} aims to facilitate the cybersecurity research and the collaboration across teams.

\end{abstract}

%
%
%
\section{Introduction}


Cybersecurity is not a product, but a process. One that needs to be continuously assessed in a periodic manner, as systems evolve and new cyber-threats are discovered. This becomes specially relevant with the re-use of (software and hardware) components and the increasing complexity of such systems \cite{bozic2017planning}. Robotics is the art of system integration \cite{mayoral2017shift}. Its modular characteristic by nature, both in hardware and software, provides unlimited flexibility to its system designers. This flexibility however comes at the cost of increased complexity. Current robotic systems are often highly complex, a condition that in most cases leads to wide attack surfaces and a variety of potential vulnerabilities, which increase the complexity of remaining secure by solely relying on the use of traditional perimeter-based approaches. 

Robot cybersecurity vulnerabilities are potential attack points in robotic systems that can lead not only to considerable loss of data, but also to safety incidents involving humans \cite{RobotHaz}. Some claim \cite{zheng2011ivda} that unresolved vulnerabilities are the main cause of loss in cyber-incidents. Given the interaction robots have with humans and our environments, the qualification of a security flaw, commonly known as \emph{triage}, seems of special relevance in the domain of robotics. Mitigating a vulnerability or a bug often requires one to first triage and reproduce the flaw in a  deterministic manner. 
Dealing with robots requires interacting with both hardware and software. The strong interdisciplinary of the robotics fields (which is an inherent characteristic of the complexity of the area) and the trade-off obtained with its modularity~\cite{Kolak2020ItTakesAVillage} demands for knowledge across multiple domains in order to comprehensively analyze a robot system. Correspondingly, triaging and reproducing security flaws in robotics can become extremely demanding in terms of time and resources, especially when cooperating with team members across multiple locations, several of which don't necessarily may have access to the same hardware. 

To address this and accelerate research, a toolbox for robot cybersecurity should facilitate the creation of virtual environments that capture not only software components, but also hardware ones. OS-virtualization and hardware-virtualization are both needed in robotics, specially with the growing adoption of modular frameworks like the Robot Operating System (ROS) \cite{quigley2009ros}. We believe that emulation and simulation should co-exist for robot cybersecurity research, and it should be possible to easily interface these with real hardware.

The present article attempts to deal with these problems by proposing a cybersecurity toolbox that aims to enable roboticists and security researchers cooperating in project to have a common, consistent and easily reproducible development environment, facilitating the security process (DevSecOps \cite{mayoral2020devsecops}) in robotics and the collaboration across multiple teams. The contributions of this work are threefold. First, we present a modular toolbox for robot cybersecurity that facilitates the cybersecurity processes. It reflects the modular nature of robotics while at the same time makes the resulting complexity easier to handle for cybersecurity researchers. Second, we demonstrate support for mixed simulation and emulation environments in robotics. This helps research  scenarios involving high-fidelity virtual versions of hardware and software robot components. Finally, through composability, the toolbox provides modules that can be selected and assembled in various combinations, providing facilities for automation of security testing, evaluation and demonstration while enhancing the triage and reproducibility of security flaws.

The remaining content is organized as follows: Section \ref{sec:background} presents previous related work. Section \ref{sec:alurity} describes the toolbox and section \ref{sec:use_cases} a series of use case experiences. Section \ref{sec:lessons_learned} presents results using the toolbox and argues about some of the lessons learned throughout the process. Finally, Section \ref{sec:conclusions} summarizes our work and draws some conclusions while hinting on future work actions.

\section{Related work}
\label{sec:background}

\setlength{\tabcolsep}{10pt}
\renewcommand{\arraystretch}{1.4}
\begin{table*}[t]
    \centering
	\resizebox{\textwidth}{!}{ 
		\begin{tabular}{|l|l|l|l|l|l|l|l|}
			\hline
			& \bf \textsc{Scope} & \bf \textsc{Simulation} & \bf \textsc{Emulation} & \bf \textsc{Mixed} & \bf \textsc{Scalable} & \bf \textsc{Flexible}  \\ 
			\hline
			DETERlab~\cite{5655108} & testbed facility, hardware testing &  \cellcolor{black!15} No  & Yes  & \cellcolor{black!15} No  & \cellcolor{black!15} Low      & Yes\\ 
			VINE~\cite{Eskridge:2015:VCE:2808475.2808486} & computer networks, dynamic and moving target defenses  & \cellcolor{black!15} No         & Yes       & \cellcolor{black!15} No  & \cellcolor{black!10} Med      & Yes\\ 
			SmallWorld~\cite{FURFARO2018791} & computer networks, network devices, hardware devices  & \cellcolor{black!15} No         & Yes       & \cellcolor{black!15} No  & \cellcolor{black!10} Med      & Yes\\ 
			BRAWL~\cite{BRAWL} & enterprise computer networks & \cellcolor{black!15} No         & Yes       & \cellcolor{black!15} No  & \cellcolor{black!10} Med      & \cellcolor{black!15} Windows \\ 
			Galaxy~\cite{schoonover2018galaxy} & computer networks & \cellcolor{black!15} No         & Yes       &  \cellcolor{black!15} No  &  \cellcolor{black!15} Low      & \cellcolor{black!15} Debian-based\\ 
			Insight~\cite{Futoransky2009SimulatingCF} & computer networks, network devices, hardware devices & Yes        & \cellcolor{black!15} No        & \cellcolor{black!15} No  & High     & Yes\\ 
			CANDLES~\cite{Rush:2015:CAN:2739482.2768429} & computer networks, game theory & Yes        & \cellcolor{black!15} No        & \cellcolor{black!15} No  & High     & Yes\\ 
			Pentesting Simulations~\cite{niculae_2018, JThesis} & computer networks, penetration testing, machine learning  &  Yes        & \cellcolor{black!15} No        & \cellcolor{black!15} No  & High     & Yes\\ 
			CyAMS~\cite{7795375} & computer networks, malware propagation & Yes & Yes  & \cellcolor{black!15} No  & High     & Yes\\ 
			CybORG~\cite{baillie2020cyborg} & computer networks, machine learning &  Yes        &  Yes       &  \cellcolor{black!15} No  &  High     & Yes\\ 
			\bf \emph{alurity} (ours) & \bf robotics, industrial devices, computer networks & \bf Yes        &  \bf Yes       &  \bf Yes  &  \bf High     &  \bf Yes\\ 
			\hline
		\end{tabular}
	}
	\caption{{A comparison of cybersecurity-focused environments and frameworks. Each one of them is evaluated according to five characteristics: simulation capabilities, emulation capabilities, mixed simulation and emulation capabilities, scalability (complex environments, multiple endpoints, etc.) and flexibility in terms of the requirements to run it. With regard the mixed column, note it refers to either the presence of both (emulation and simulation) as complete entities in an scenario, or alternatively the possibility to simulate some parts, and emulate others both together in a combined entity.}}
	\label{table:previous_work}
\end{table*}

The mitigation and patching of vulnerabilities has been an active area of research \cite{ma2001sharing, ALHAZMI2007219, Shin2011Vulnerabilities, Finifter2013BugBounty, McQueen2009Zeroday, Bilge:2012:BWK:2382196.2382284} in computer science and other technological domains. Unfortunately, with robotics being an interdisciplinary field composed from a set of heterogeneous disciplines (including computer science), to the best of our knowledge and literature review, except for a few preliminary tools and reviews \cite{vilches2019introducing, RVSS, ekenna2020clustering, mayoral2020can, dieber2020penetration, lacava2020current}, not much vulnerability mitigation research related to robotics has been presented so far.

Due to their expensive prices and shortage, getting access to robots for testing is somewhat comparable to the difficulties researchers face when analyzing complex computer networking systems. The Galaxy framework~\cite{schoonover2018galaxy} researched this and proposed a high-fidelity computer network emulation tool designed to support rapid, parallel experimentation with the automated design of software agents in mind. By leveraging virtualization, the authors of Galaxy create a high-fidelity clone of a network that takes snapshots to always have a copy of this known base state. With this cloned network, researchers can distribute the work and perform experiments to see how the network performs with certain kind of attacks. 

Virtualization is a commonly used technique in other areas allowing developers and security practitioners to more easily test, reuse and ship systems. When using virtualization, there are often two approaches, emulation and simulation. Emulation is generally implemented through hardware-virtualization, typically using full-virtualization with type two hypervisors (e.g. VirtualBox, VMWare Worksation or QEMU) \cite{grattafiori2016understanding}, commonly referred to as Virtual Machines (VMs). This provides complete isolation between guest kernels and host, while allowing to run many different operating Systems (OS) within the same physical host. Simulation is often implemented with simpler abstractions such as state machines or through OS-virtualization, using a shared kernel across both the host and guests. OS-virtualization is typically referred to as "containers" in Linux \cite{grattafiori2016understanding}, and is widely considered the most efficient virtualization method for its highest performance and fastest "start-up" time. Different approaches to simulation are presented at \cite{Futoransky2009SimulatingCF, Rush:2015:CAN:2739482.2768429, niculae_2018, JThesis, 7795375, baillie2020cyborg}.

The authors of \cite{schoonover2018galaxy} confirm that shifting from simulated to emulated experiments removed the luxury of real-world abstractions and acknowledge that VMs must be power cycled, or reverted to reset states, as opposed to quicker simulations. They further mention that VM emulation is bulky and computationally expensive, whereas containers (OS-virtualization) use fewer system resources and have faster power cycle actions. A concerning matter regarding OS-virtualization is that unlike VMs, which emulate the entire network architecture stack, containers share the host kernel which comes at the cost of fidelity. This was raised by \cite{schoonover2018galaxy} however Handigol et al. \cite{handigol2012reproducible} and Heller \cite{heller2013reproducible} demonstrated how containers, could be used without a relevant loss of fidelity, which further encourages the use of containerization. Other approaches to emulation include \cite{5655108, Eskridge:2015:VCE:2808475.2808486, FURFARO2018791, BRAWL, 7795375} or \cite{baillie2020cyborg}.

A similar experience to \cite{schoonover2018galaxy} was described more recently by \cite{baillie2020cyborg} which targets research in Autonomous Cyber Operations (ACO). The authors of \cite{baillie2020cyborg} proposed a high-fidelity emulation using VMs, and a low-fidelity simulation using a finite state machine design that could model networks. Virtualization of the different scenarios was realized by using a description file (YAML) and a flag determined the use of either simulation or emulation. The results obtained of this dual approach show promise and the authors already hint that future work might train in low-fidelity simulation, and later optimize in high-fidelity emulation.


Our work builds on top of these past experiences and proposes a toolbox that facilitates the virtualization of robotic scenarios with the capability of doing both high-fidelity emulation and simulation using VMs and containers, respectively. Moreover, by extending the networking abstractions of both VMs and containers, our toolbox allows for mixed heterogeneous virtual environments, where emulated endpoints (VMs) can interact with simulated ones (containers), providing more flexibility for roboticists and security researchers. Table \ref{table:previous_work} presents our approach compared to past work.

\section{alurity}
\label{sec:alurity}

\begin{figure*}[!t]
    \includegraphics[width=0.9\textwidth]{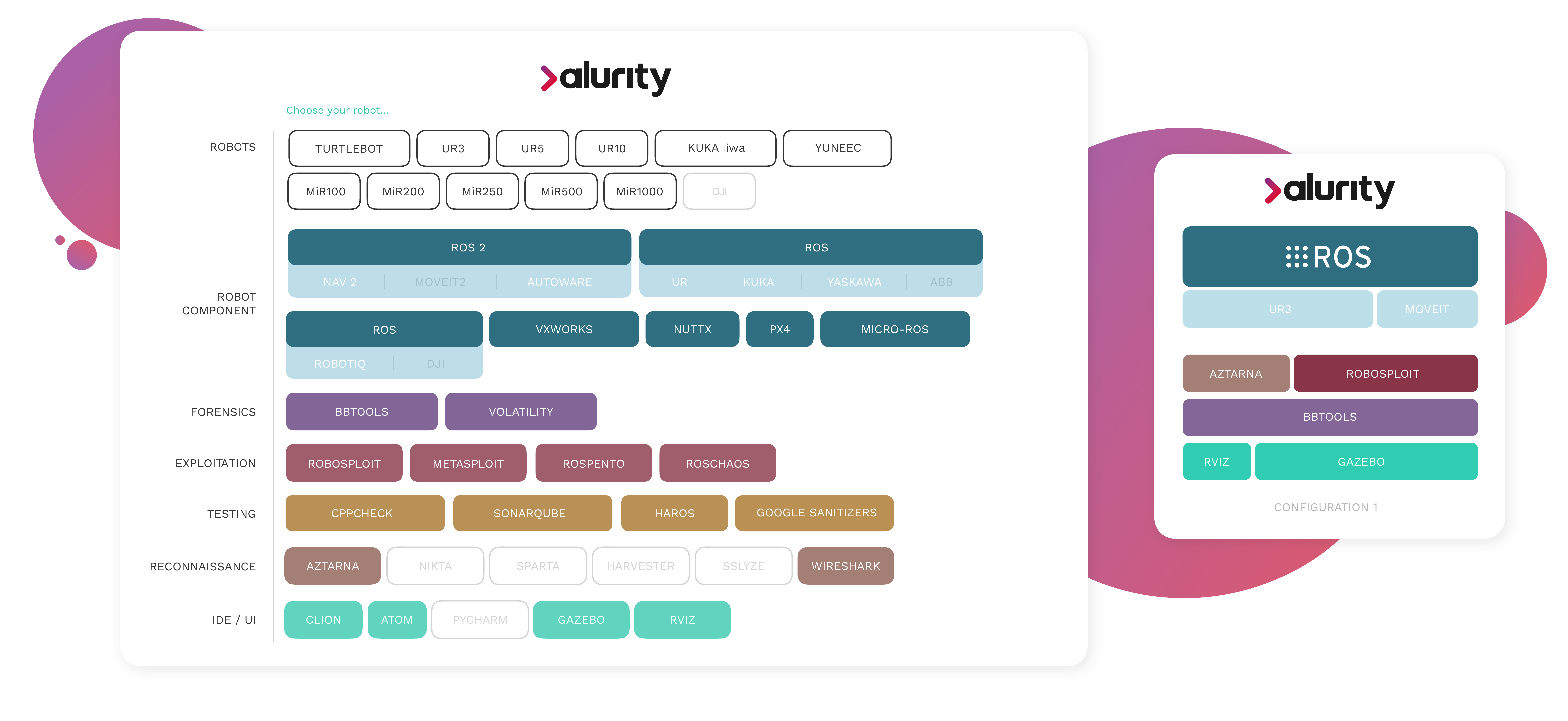}
    \centering
    \caption{
        \footnotesize \textbf{\emph{alurity} toolbox for robot cybersecurity}. The toolbox classifies tools in seven groups that favour composition and re-usability among them. The groups are: robots, robot components, forensics, exploitation, testing, reconnaissance and IDE/UI. In the left, we depict a subset of the tools available, across the different groups. The right hand side illustrates a composition of tools into a construction that will be used during the robot cybersecurity research effort. The subject constructed features a base ROS \cite{quigley2009ros} module, and a number of submodules composed below including the UR3 ROS driver, the MoveIt \cite{chitta2012moveit} Motion Planning Framework ROS packages, the aztarna \cite{mayoral2018aztarna} footprinting tool for robots, the Gazebo \cite{koenig2004design} robot simulator or robosploit, a proprietary exploitation framework for robots, among others.
    }
    \label{fig:toolbox}
\end{figure*}

\emph{Alurity} is a modular and composable toolbox for robot cybersecurity. It ensures that both roboticists and security researchers working on a project, have a common, consistent and easily reproducible development environment facilitating the security process and the collaboration across teams. It's available for Linux (across distributions) and Mac OS, with limited support at the time of writing for Windows. Featuring dozens of different tools in the form of individual components, \emph{alurity} simplifies and speeds up the cybersecurity research in robotics. The toolbox organizes security and robotics components in different groups that allow for easy composition and use. Figure \ref{fig:toolbox} illustrates this aspect and the different groups, while constructing a robotic subject with it.

Similar to \cite{baillie2020cyborg}, \emph{alurity} toolbox allows to build scenarios using a simple description file in YAML format. Listing \ref{alurity:example} provides an example of such format while modeling the scenario of Figure \ref{fig:use_case:ur}.

\lstset{morekeywords={network, networks, name, driver, internal, encryption, subnet, ip, firewalls, container, ingress, egress, rules, modules, base, cpus, memory, mount, volume}}
\lstset{label={alurity:example}}
\lstset{basicstyle=\tiny,
    numbers=none,
    firstnumber=1,
    stepnumber=1}
\lstset{caption={
    \footnotesize \emph{Alurity} YAML syntax to simulate a UR3 controller and an attacker, as part of the use case depicted in Figure \ref{fig:use_case:ur}.}
}
\begin{lstlisting}
networks:
  - network:
    - name: process-network
    - driver: overlay
    - internal: true
    - encryption: false
    - subnet: 12.0.0.0/24

  - network:
    - name: cloud-network
    - driver: overlay
    - subnet: 17.0.0.0/24
    
containers:
  - container:
    - name: "ur3"
    - modules:
         - base: registry.gitlab.com/aliasrobotics/offensive/alurity/robo_ur_cb3_1:3.13.0
         - network:
           - process-network
    - ip: 12.0.0.20  # assign manually an ip address
    - cpus: 4
    - memory: 2048

  - container:
    - name: attacker
    - modules:
         - base: registry.gitlab.com/aliasrobotics/offensive/alurity/comp_ros:melodic-scenario
         - volume: registry.gitlab.com/aliasrobotics/offensive/alurity/expl_robosploit/expl_robosploit:latest
         - volume: registry.gitlab.com/aliasrobotics/offensive/alurity/reco_aztarna:latest
         - volume: registry.gitlab.com/aliasrobotics/offensive/alurity/deve_gazebo:latest
         ...
         - network:
           - process-network
           - cloud-network
    - extra-options: ALL

\end{lstlisting}

\subsection{Simulated, emulated or real endpoints}

Dealing with robots requires interacting with both hardware and software. Correspondingly, a toolbox that helps security researchers and roboticists in their cybersecurity efforts should consider virtual environments that capture not only software components, but also hardware ones. OS-virtualization and hardware-virtualization are both needed. In other words, emulation and simulation should co-exist. 

Virtualization in \emph{alurity} defaults to simulation. Built using containers (OS-virtualization) it delivers an efficient, high-fidelity, high-performance and fast virtualization method. Most \emph{alurity} modules (tools) are containerized this way and can be easily composed as demonstrated in listing \ref{alurity:example}. High-fidelity emulation is implemented using VMs (hardware-virtualization)---more specifically---on top of VirtualBox VMs (full-virtualization, type two hypervisor). The YAML syntax for emulation is similar to the one for simulation and exemplified in listing \ref{alurity:example2}:

\lstset{morekeywords={network, networks, name, driver, internal, encryption, subnet, ip, firewalls, container, ingress, egress, rules, modules, base, cpus, memory, mount, volume}}
\lstset{label={alurity:example2}}
\lstset{basicstyle=\tiny,
    numbers=none,
    firstnumber=1,
    stepnumber=1}
\lstset{caption={
    \footnotesize \emph{Alurity} YAML syntax to emulate an ABB’s IRC5 Compact robot controller as per the use case depicted in Figure \ref{fig:use_case:abb}.}
}
\begin{lstlisting}
vms:
  - vm:
    - name: irc5
    - path: $\textdollar$(pwd)/vms/irc5
    - network: process-network
    - ip: 12.0.0.100
    - cpus: 2
    - memory: 2048
\end{lstlisting}

\emph{Alurity} allows to easily build robotic subjects of study through the virtualization of different robots, networks (e.g. LANs, serial comms, VPNs, VLANs, VXLANs, WLANs, etc) and robot components, including security protection mechanisms as depicted in Figure \ref{fig:use_case:kics}. It supports both, simulation and emulation of modules and allows for mixed environments where emulated endpoints can co-exist and interact (networking-wise) with simulated ones. This logic is implemented at the core of the toolbox through private network internals to the host so that containers on this network can communicate with VMs. Behind the scenes, \emph{alurity} creates the necessary bridges, internal virtual network interfaces, iptables rules, and host routes to make this connectivity possible. 

Beyond virtualization (emulation or simulation), \emph{alurity} was built to include real robot endpoints in the scenarios. Correspondingly, the toolbox and its networking capabilities allow to interface with public networks and interoperate with real robots, while leveraging the power to build virtual attackers and weaponize them with different tools and exploits.

\subsection{Flows}

\emph{Alurity} operates mostly through its Command Line Interface (CLI), which facilitates to download your configuration YAML file modules and start the scenarios, to begin with the corresponding experiments. Beyond core modules, \emph{alurity} provides an additional capability called \texttt{alurity flows} meant to facilitate evidence generation and reproduction. Flows are series of commands within a scenario built with \emph{alurity}. A flow allows to automate interactions between endpoints through the launch of a series of commands and applications via simple YAML format. Simply put, flows allow researchers to reproduce each other's work and results by simply providing a YAML file with the scenario description and one (or multiple) flows. \emph{alurity} flows are built using a mix of tools including byobu, tmux and other utilities which are embed into a base layer present in every \emph{alurity} module. Listing \ref{alurity:example3} exemplifies a flow where an attacker targets the ROS core of an industrial setup. 

To further automate security research, \emph{alurity} YAML files, including the flows, can be integrated into vulnerability tickets of robot vulnerability databases, such as the Robot Vulnerability Database (RVD) \cite{vilches2019introducing}, allowing other researchers to simply \texttt{alurity run --rvd <ticket-id>} fetching the corresponding \emph{alurity} YAML portion and launching its flow for reproducing the vulnerability. Examples of this include \cite{rvd2554, rvd2555, rvd2556} or \cite{rvd2558} among others.

\lstset{morekeywords={network, networks, name, driver, internal, encryption, subnet, ip, firewalls, container, ingress, egress, rules, modules, base, cpus, memory, mount, volume, window, commands, command, split, flow}}
\lstset{label={alurity:example3}}
\lstset{basicstyle=\tiny,
    numbers=none,
    firstnumber=1,
    stepnumber=1}
\lstset{caption={
    \footnotesize \emph{Alurity} YAML syntax to create a flow that launches a FIN-ACK attack against the ROS master machine ($S_7$) in the use case depicted in Figure \ref{fig:use_case:rosi}}
}
\begin{lstlisting}
flow:
  # rosmachine
  - container:
    - name: rosmachine
    - window:
      - name: ros
      - commands:
        - command: "source /opt/ros/melodic/setup.bash"
        - command: "roscore"
        - split: horizontal
        - command: "source /opt/ros/melodic/setup.bash"
        - command: "sleep 10"
        - command: "rostopic echo /chatter"
        - split: horizontal
        - command: "source /opt/ros/melodic/setup.bash"
        - command: "sleep 10"
        - command: "rostopic hz /chatter"
        
  # attacker
  - container:
    - name: attacker
    - window:
      - name: setup
      - commands:
        - command: "wireshark -i eth0 . &"
        - split: horizontal
        - command: "apt-get update && apt-get install -y tcpdump iptables"
    - window:
      - name: attack
      - commands:
        - command: "source /opt/ros/melodic/setup.bash"
        - command: 'export ROS_MASTER_URI="http://12.0.0.2:11311"'
        - command: "cd /home/alias"
        - command: "sleep 10"  # wait until roscore is ready
        - command: "/opt/ros/melodic/lib/roscpp_tutorials/talker"
        - split: horizontal. % create a terminal split and spawns a new shell
        - command: "sleep 10"  # wait until tools have been installed and roscore
        - command: "source /opt/ros/melodic/setup.bash"
        - command: 'export ROS_MASTER_URI="http://12.0.0.2:11311"'
        - command: "cd /home/alias"
        - command: "iptables -I OUTPUT -s 12.0.0.4 -p tcp --tcp-flags RST RST -j DROP"
        - command: "iptables -I OUTPUT -s 12.0.0.4 -p tcp --tcp-flags FIN FIN -j DROP"
        - command: 'python3 fin_ack_dos.py'
        - command: 'robosploit -m exploits/ros/fin_ack -s "target 12.0.0.2"' 
    - select: attack
\end{lstlisting}

\subsection{Pipelines}

When performing a security exercise, researchers often spend a considerable amount of time with information gathering and testing, for bug identification. This process is often performed manually, consuming resources that could instead be used for exploitation and/or mitigation. In order to alleviate this effort we built \texttt{alurity pipelines}, automated end-to-end and on-demand security pipelines that automatically build an \emph{alurity} YAML file with selected tools and "runs them" through the desired target module. For each security flaw identified, the pipeline generates a YAML file describing the flaw using RVD \cite{vilches2019introducing} taxonomy\footnote{Available at \url{https://bit.ly/36Y7uxt}} either in a local directory or pushing the flaw directly as an issue to Github or Gitlab. This way, pipelines support the robotics DevSecOps \cite{mayoral2020devsecops} pushing each potential vulnerability to a centralized flaw management registry.

\section{Use cases}
\label{sec:use_cases}

\begin{figure*}[!h]
    \centering
    \begin{subfigure}[b]{0.49\textwidth}
        \includegraphics[width=0.8\textwidth]{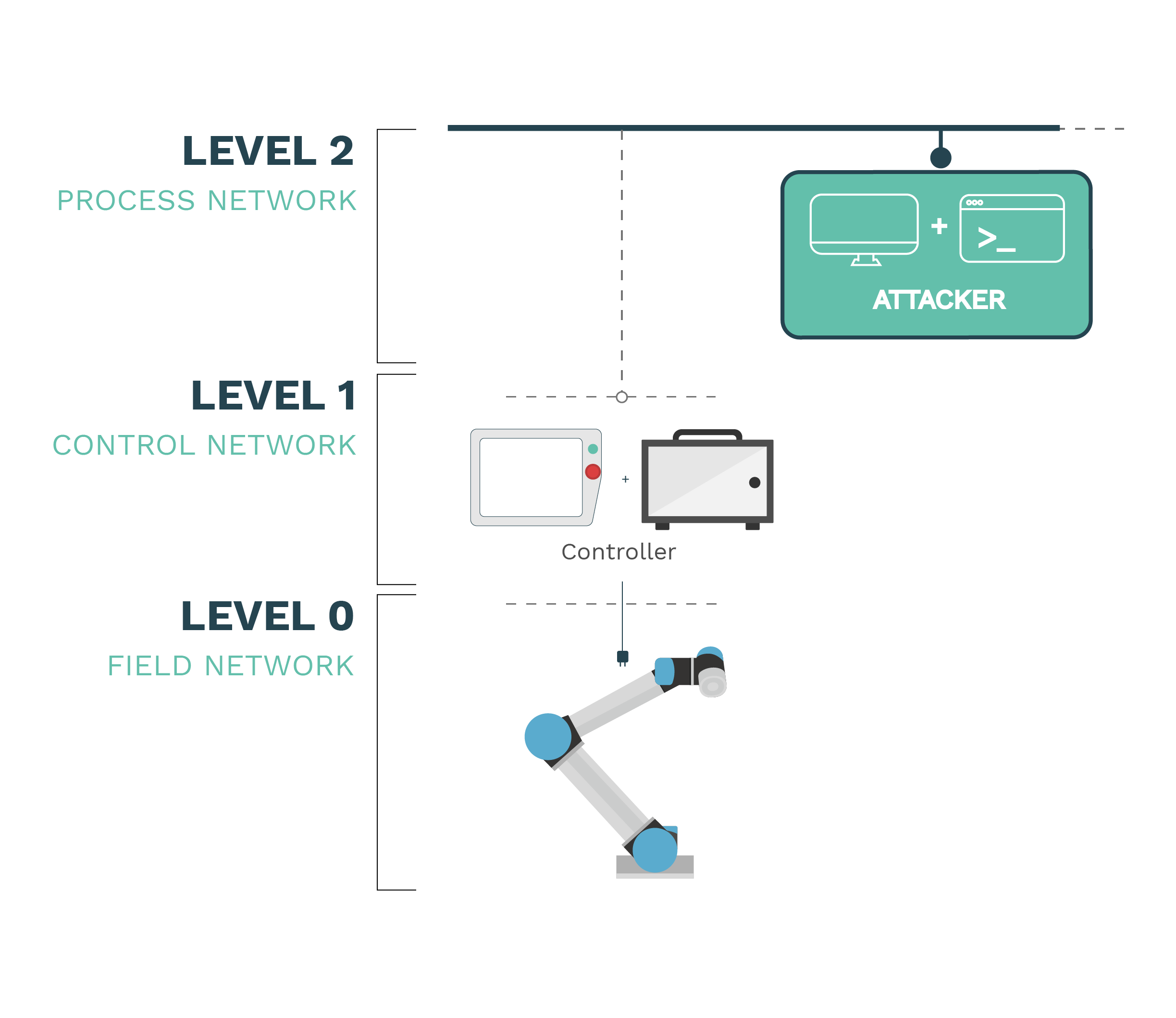}
        \caption{
        \footnotesize Simulated scenario with a Universal Robots UR3 robot and a CB3.1 controller connected to a process level network where an attacker has compromised a machine.
        }
        \label{fig:use_case:ur}
    \end{subfigure}
    ~ 
    \begin{subfigure}[b]{0.49\textwidth}
        \includegraphics[width=0.8\textwidth]{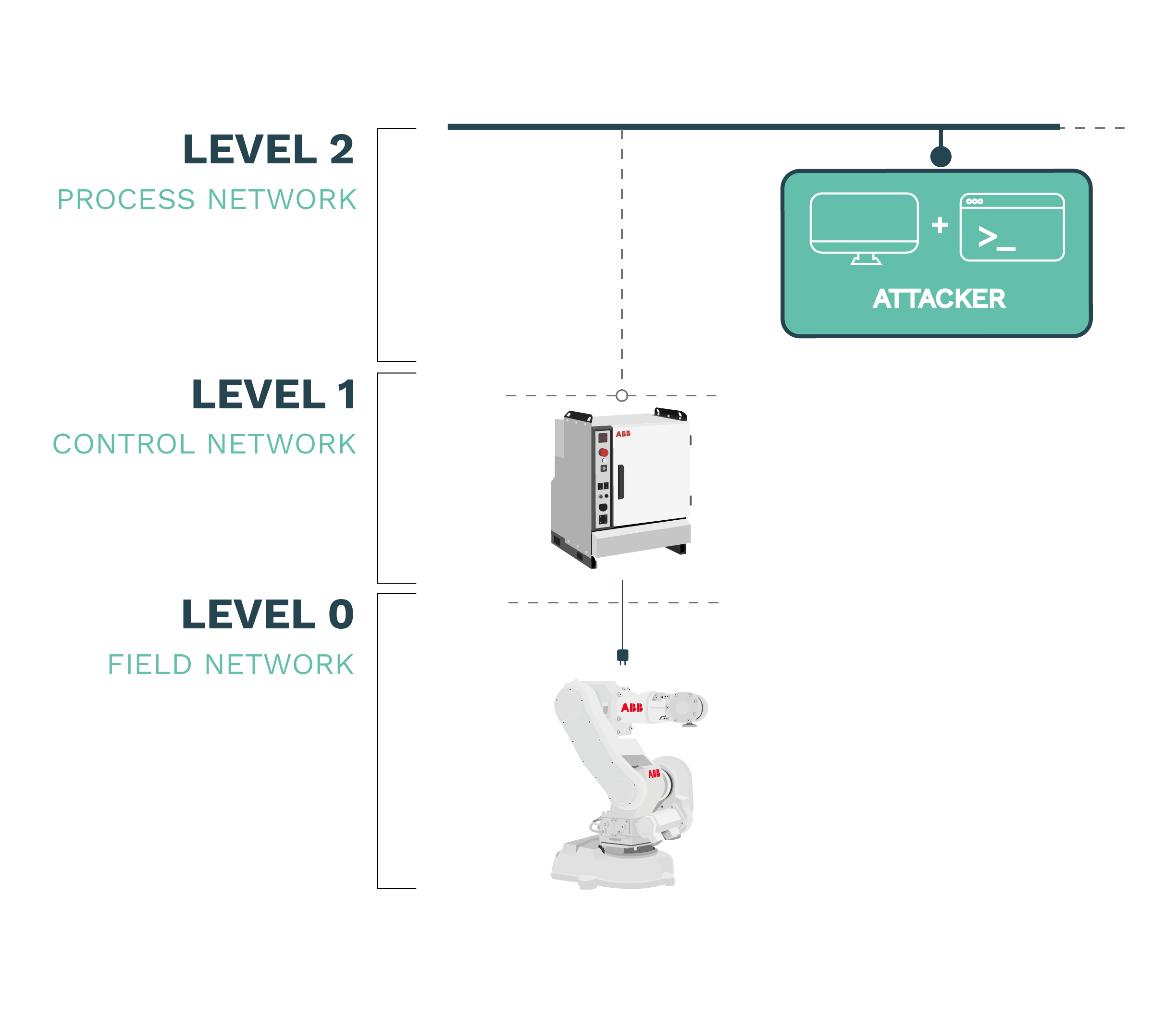}
        \caption{
        \footnotesize Mixed virtual and real scenario composed by an ABB’s IRC5 compact robot controller with the IRB140 Manipulator and the DSQC 679 teach pendant. The controller is assumed connected to a process level network where an attacker has compromised a machine.
        }
        \label{fig:use_case:abb}
    \end{subfigure}
    ~
    \begin{subfigure}[b]{0.49\textwidth}
        \includegraphics[width=1\textwidth]{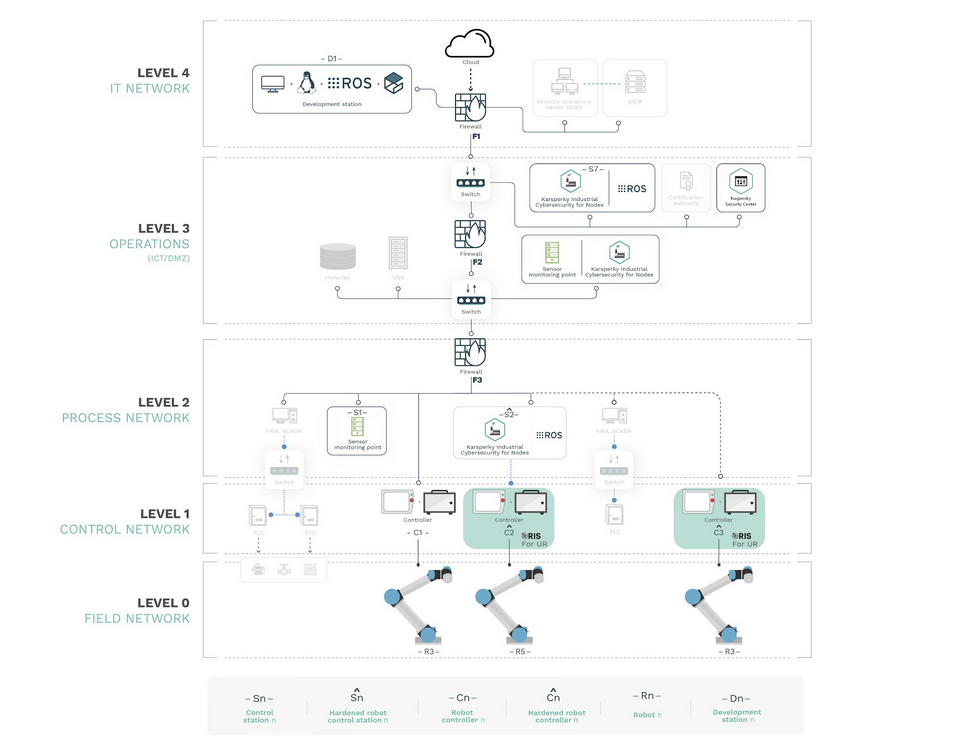}
        \caption{
        \footnotesize Mixed virtual industrial factory automation process with security integrated, namely the Kaspersky Industrial CyberSecurity (KICS) solution and Alias Robotics' Robot Immune System (RIS).
        }
        \label{fig:use_case:kics}
    \end{subfigure}
    ~ 
    \begin{subfigure}[b]{0.49\textwidth}
        \includegraphics[width=1\textwidth]{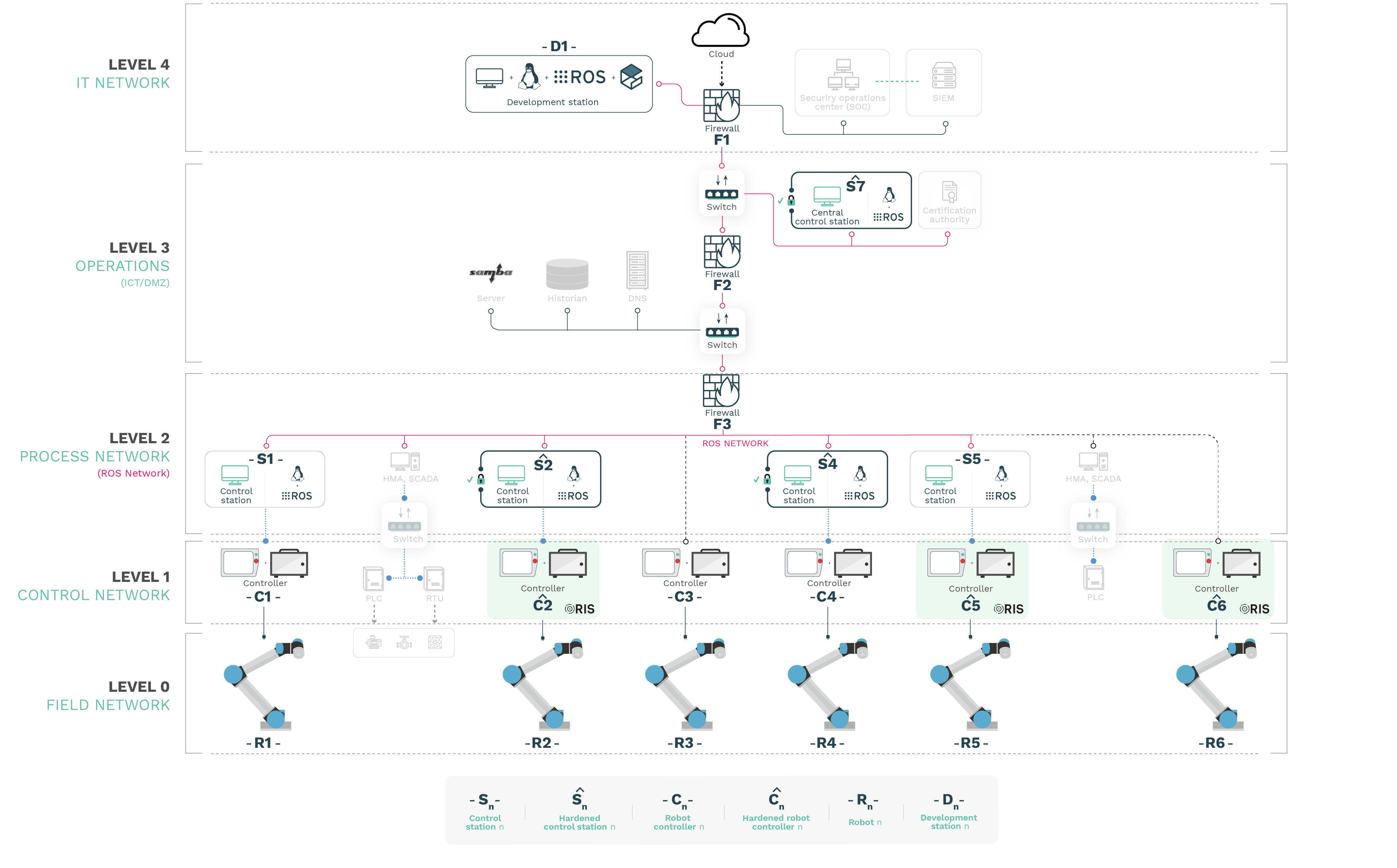}
        \caption{
        \footnotesize A simulated synthetic industrial scenario powered by ROS and ROS-I packages which presents a network segmented in 5 levels with segregation implemented following security standards.
        }
        \label{fig:use_case:rosi}
    \end{subfigure}
    
    \caption{Use cases realized with \emph{alurity} toolbox.}
    \label{fig:use_case}
\end{figure*}



To illustrate the flexibility of the \emph{alurity} toolbox, Figure \ref{fig:use_case} depicts several use cases of robot cybersecurity research conducted with it. For each one of these scenarios, either simulation and/or emulation was used, interfacing when appropriate with real hardware for final validation of the flaws. We discuss below briefly each use case and the virtualization mechanisms used:

\subsection{Penetration testing Universal Robots's UR series}

Using \emph{alurity}, we conducted a penetration testing security exercise with the aim of providing a practical assessment of the UR3, UR5 and UR10 robot insecurities. Before the exercise, there were five reported vulnerabilities for these collaborative robots. After the penetration testing assessment, more than 80 were publicly available, many with proof-of-concept exploits.

Figure \ref{fig:use_case:ur} depicts the scenario we assumed for the exercise  with a Universal Robots UR3 robot and a CB3.1 controller connected to a process level network, where an attacker had compromised a machine. The scenario was simulated with \emph{alurity} and both the robot controller and the attacker were containerized. The mechanics of the robot were ignored for the purpose of the study. While emulation could have added additional fidelity to the robot controller virtualization at the expense of additional time, we decided not to proceed in that direction and instead focused mostly on simulation, with the exception of physical attack vectors, which were conducted against the real robot with an \emph{alurity} simulated attacker. Public results were discussed at \cite{pentestingur}. A complete list of the vulnerabilities released is available at \url{https://bit.ly/3183cQF}.


\subsection{Penetration testing ABB's IRC5 controller}

We studied ABB’s IRC5 compact robot controller with the IRB 140 manipulator and the DSQC 679 teach pendant. For the use case, we assumed the controller was connected to a simulated process level network where an attacker had compromised a machine. Given the closed environment of ABB's technology, in this case, we used \emph{alurity} to mix simulation, emulation and real hardware while conducting the penetration testing assessment. The Attacker was simulated, the manipulator and the teach pendant were real and the controller was either real or emulated. Emulations of the controller consisted of isolated components extracted from its firmware. Public results of the exercise were discussed at \cite{pentestingabb}. A complete list of the vulnerabilities released during the exercise conducted within 2020 for ABB is available at \url{https://bit.ly/2FsZQzX}

\subsection{Penetration testing KICS}

This use case considered a more sophisticated environment involving more elements and capturing a selected portion of a secure industrial factory automation process. In particular, we focused our study on how insecure robots could be used as entry points to further compromise the Kaspersky Industrial CyberSecurity (KICS) solution, a commercial product. KICS imposes strong demands on the kernel and the hardware devices available from the file system, thereby after struggling with containerization for a while, we opted for emulating KICS-related entities including KSC, KICS for Networks  and the associated sensors\footnote{Details on KICS solution is beyond the scope of this article.}. We used \emph{alurity} to harmonize all networking aspects of the scenario and simulate the rest of the endpoints involved including robot controllers, workstations and other ICS elements. Results of the exercise are not publicly available at the time of writing.

\subsection{Red teaming ROS-Industrial}
In this use case we aimed to answer the question of whether ROS could be used securely for industrial use cases. We did so experimentally by performing a targeted offensive security exercise in a use case involving ROS-Industrial and ROS packages. We built a synthetic industrial scenario which presented a network segmented in 5 levels, with segregation implemented following standardized recommendations. \emph{Alurity} was used to simulate all entities and network elements, including those necessary for segmentation and segregation. By doing so we obtained a lightweight yet high-fidelity environment whereto perform attacks. Results of the exercise are discussed at \cite{redteamingrosi} and further documented in \cite{redteamingrosindustrial_whitepaper, mayoral2020can}.

\section{Lessons learned and results}
\label{sec:lessons_learned}


At the time of writing \emph{alurity} toolbox features more than 70 different modules, including security tools, industrial common endpoints (e.g. PLCs or HMIs), robots and robot components. Due to our aim to use well-tested and established technology, as opposed to reinvent the wheel on security tooling, we re-used as much past work as possible when building the toolbox. Specially on the simulation and virtualization infrastructure side which led us to save a significant amount of resources, focusing our efforts instead on a) usability and b) optimizing the delivery of cybersecurity research products, namely vulnerabilities, proof-of-concept (PoC) exploits and mitigations. Maintaining this focused attitude on re-use is what led us to produce usable cybersecurity research results. Out of the use cases in Figure \ref{fig:use_case}, we produced several dozens of vulnerabilities and received multiple new CVE IDs while responsibly disclosing the flaws to the upstream manufacturers. Public disclosure of these vulnerabilities happened in RVD \cite{vilches2019introducing} with results catalogued by manufacturer, robot or robot component\footnote{Refer to section \ref{sec:use_cases} for examples of which vulnerabilities were found}.

Thanks to the use of virtualization, a significant amount of time was saved. Research could run in parallel among researchers and across groups, with the few physical robots available used only for final validation. The use of high-fidelity simulation also accelerated the process of dynamic testing, wherein the robot often entered into unstable states. Power-booting simulations required just a few seconds (as opposed to minutes with emulated robots, or an even longer time if having to re-flash firmware in the real robot). A particular \emph{alurity} feature that simplified peer-triage was \texttt{alurity flows} which when embed into vulnerability reports (as in \cite{rvd2554} or \cite{rvd2555}) complemented nicely exploit PoCs with additional context. Another aspect of the toolbox that proved useful was to run known attacks on new firmware versions of the robots or robot components. Specially in combination with simulations described by the YAML syntax which allowed for quick changing of the targets by simply modifying only one YAML line. The same applied to new security mechanisms, which could be tested against attacks, and across robots easily with \emph{alurity}.

Overall, after these experiences and although we don't feel we could generalize for all use cases, we strongly recommend security researchers to consider testing and triaging in simulation before defaulting to emulation and ultimately, to real endpoints. This will save them a considerable amount of time and will facilitate reproduction and interactions with upstream vendors, who are likely the final recipients of their security results.








\section{Conclusions}
\label{sec:conclusions}






In this study we presented \emph{alurity}, a modular and composable toolbox for robot cybersecurity that ensures that both roboticists and security researchers have a common, consistent and easily reproducible development environment. We described the architecture of \emph{alurity} including its capabilities for virtualization. We also described additional features that facilitate the security process and cooperation across teams, namely flows and pipelines. We then presented four real use cases conducted with \emph{alurity} and shared some lessons learned, emphasizing that simulation appears to speed up significantly cybersecurity research in robotics.

The lack of documentation for \emph{alurity} is currently the biggest hurdle  preventing us from disclosing it to a wider audience. Future work should focus on this. We also plan to look into preserving the state of the simulated scenarios across runs, as well as improved Windows support.


\section*{ACKNOWLEDGMENTS}

\noindent This research has been funded by Alias Robotics, by the INNOEM-2020/00037 Álava Innova research grant throughout the Economic Development and Innovation Service of the Foral Council of Álava and finally, also by the Spanish Government through CDTI Neotec actions (SNEO-20181238).  Thanks also to the City Council of Vitoria-Gasteiz and the Basque Government, throughout the Business Development Agency of the Basque Country (SPRI). Special thanks to BIC Araba and the Basque Cybersecurity Centre (BCSC) for the support provided. \\


\bibliographystyle{IEEEtran}
\bibliography{IEEEabrv,bibliography}



\end{document}